\documentclass[preprint,12pt]{elsarticle}

\usepackage{amssymb}

\usepackage{subcaption}
\usepackage{pifont}
\usepackage{amsmath}
\usepackage{amssymb}
\usepackage{booktabs}
\usepackage{multirow}
\usepackage{array}
\usepackage{algorithm}
\usepackage{algorithmicx}
\usepackage{algpseudocode}
\usepackage{xcolor}
\usepackage{hyperref}

\begin{document}

\begin{frontmatter}



\title{POTLoc: Pseudo-Label Oriented Transformer for \\ Point-Supervised Temporal Action Localization}


\author[inst1]{Elahe Vahdani}

\author[inst1]{Yingli Tian}

\affiliation[inst1]{organization={The City College},
            addressline={City University of New York}, 
            city={New York},
            postcode={10031}, 
            state={NY},
            country={USA}}


\begin{abstract}
This paper tackles the challenge of point-supervised temporal action detection, wherein only a single frame is annotated for each action instance in the training set. Most of the current methods, hindered by the sparse nature of annotated points, struggle to effectively represent the continuous structure of actions or the inherent temporal and semantic dependencies within action instances. Consequently, these methods frequently learn merely the most distinctive segments of actions, leading to the creation of incomplete action proposals. This paper proposes POTLoc, a \textbf{P}seudo-label \textbf{O}riented \textbf{T}ransformer for weakly-supervised Action \textbf{Loc}alization utilizing only point-level annotation. POTLoc is designed to identify and track continuous action structures via a self-training strategy. The base model begins by generating action proposals solely with point-level supervision. These proposals undergo refinement and regression to enhance the precision of the estimated action boundaries, which subsequently results in the production of `pseudo-labels' to serve as supplementary supervisory signals.
The architecture of the model integrates a transformer with a temporal feature pyramid to capture video snippet dependencies and model actions of varying duration. The pseudo-labels, providing information about the coarse locations and boundaries of actions, assist in guiding the transformer for enhanced learning of action dynamics. POTLoc outperforms the state-of-the-art point-supervised methods on THUMOS'14 and ActivityNet-v1.2 datasets. Our code is available at \href{https://github.com/elahevahd/POTLoc}{https://github.com/elahevahd/POTLoc}.
\end{abstract}


\end{frontmatter}


\section{Introduction}



Automated video analysis is attracting substantial attention in the realm of computer vision and multimedia applications, largely due to its potential utility across various fields \cite{he2018anomaly, cioppa2020context, giancola2018soccernet, rasouli2019autonomous, mahadevan2019av, yao2020and}. A central task in this arena is Temporal Action Localization (TAL) in untrimmed videos, which aims to detect the temporal boundaries of actions and identify their categories \cite{vahdani2023deep}. Although recent fully-supervised TAL methods \cite{zhang2022actionformer,wang2023videomae, shi2023tridet} have demonstrated significant progress, they necessitate time-consuming and expensive annotation of temporal boundaries and action labels for each action instance in training videos. To circumvent the requirement for exhaustive annotations, many researchers are gravitating towards the development of weakly-supervised models, which only mandate a minimal set of ground-truth annotations, such as video-level labels. Nonetheless, weakly-supervised models typically lag behind their fully-supervised counterparts in terms of performance, primarily due to limited annotations and the models' constrained capacity to comprehend and learn the structure of actions. To mitigate this performance disparity, the notion of point-level supervision has been introduced \cite{moltisanti2019action, ma2020sf, lee2021learning, yang2021background}. This approach entails annotating a single frame within the temporal window of each action instance in the input video. Even though point-level supervision demands slightly more annotations than weak supervision, it substantially reduces the labeling costs compared to full supervision. Additionally, it imparts vital information about the coarse locations and the overall count of action instances, thereby enriching the model's grasp of action structures.

Due to the sparse nature of annotations in point-level supervision, existing methods frequently fail to effectively model the continuous structure of actions. Prior efforts to augment annotations have involved the generation of pseudo action and background frames, as highlighted in several studies \cite{ma2020sf, ju2020point, lee2021learning, yang2021background}. These pseudo-labeled frames contribute additional supervisory signals to the model, thereby improving its capacity to discern actions from the background. However, in the majority of these approaches, the pseudo-labeled frames are either discontinuous or they cover only fragments of the action intervals. Consequently, they often learn just the most distinctive portions of actions, which ultimately results in the production of incomplete action proposals. To counteract this issue, Lee \textit{et al.}, in \cite{lee2021learning}, developed a framework that employs an action-background contrast method to better understand action completeness, thereby fostering a more comprehensive understanding of action sequences. However, this model still falls short in adequately representing temporal dependencies within actions.

In this paper, we introduce a point-supervised framework that is designed to capture the continuous structures of actions, even in the face of extremely sparse point-level annotations. Training with only point-level supervision, the base model initially generates noisy action proposals for the training set. These action proposals are subsequently refined and adjusted to generate ``pseudo-labels" on the training set using our proposed algorithm. The pseudo-labels represent estimated temporal intervals surrounding the annotated points and are likely to align with action instances. Our pseudo-label generation algorithm is designed to discard the proposals that are potentially redundant, and to adjust those that are either excessively long or overly short. For each annotated point within the training set, we retain only the highest-scoring proposal and adjust its boundaries based on the statistics of the proposals. Importantly, our pseudo-label generation algorithm relies solely on the point-level labels and statistics of the generated proposals. Apart from the given annotated points, no ground-truth labels are used in this step. The generated pseudo-labels act as additional supervisory signals to guide our POTLoc model. 

To fully leverage the rich information provided by the pseudo-labels, POTLoc integrates a transformer with a temporal feature pyramid, effectively employing a multi-scale temporal transformer. Training multi-scale transformers for action detection under weak supervision is underexplored due to the scarcity of annotated frames. Our framework shows that we can proficiently train a transformer backbone with sparse point-level annotations. Our transformer utilizes local self-attention, aiding in the modeling of temporal dependencies within video snippets and learning the structure of actions. The temporal feature pyramid facilitates modeling actions of varying duration. The pyramid's lower levels are optimal for detecting shorter actions, while the higher levels, with their larger receptive fields, are suited for modeling longer actions. The pseudo-labels provide information about the coarse location and boundaries of actions, which aids in better guiding our multi-scale temporal transformer to learn action dynamics. Three loss functions are employed to optimize the model to effectively distinguish actions from background and accurately classify different action classes. 

We incorporate a sampling strategy during training to select the frames around the annotated points within a radius parameter and inside the boundaries of pseudo-labels, driven by two primary motivations. First, this sampling method selects snippets that are closer to the annotated points (more indicative of the action) while avoiding farther snippets (action boundaries) that can be ambiguous or contain transitional movements not representative of the action. Our experiments show that the pseudo-labels sampling improves the performance. Second, the sampling helps mitigate the issue of training the model with false positives (background frames incorrectly predicted as actions), which are more likely to exist within the boundaries of pseudo-labels. The main contributions of our work are outlined below.

\begin{itemize}

    \item We propose an innovative point-supervised framework (POTLoc) to capture the continuous structures of actions, despite relying solely on sparse point-labels.

    \item We design a novel self-training strategy to generate supplementary supervisory signals (i.e. pseudo-labels) for point-supervised action localization. This is accomplished by refining and adjusting the noisy action proposals, which are predicted by a base point-supervised model on the training set. This procedure is based on analyzing the statistics of the action proposals and their locations in relation to the annotated points.

    \item Our self-training approach enables the training of a multi-scale transformer backbone with limited supervision. The task of training transformers for action detection under weak supervision was previously underexplored, due to the large number of parameters and the scarcity of annotated frames. The multi-scale temporal transformer, guided by the generated pseudo-labels, learns to model the dependencies of video snippets and actions of varying duration.

    \item We incorporate a pseudo-labels sampling strategy to mitigate the issue of training the model with false positives and to train the model with more representative snippets.
    
           

    \item POTLoc surpasses the state-of-the-art point-supervised methods on THUMOS'14 and ActivityNet-v1.2 datasets.

\end{itemize}

\section{Related Work}

\textbf{Fully-supervised TAL.} Fully-supervised methods are categorized into anchor-based and anchor-free. Anchor-based methods generate dense proposals, distributed across temporal locations \cite{gao2017turn,gao2017cascaded,chao2018rethinking}. Anchor-free methods employ a bottom-up grouping strategy to generate proposals with precise boundaries and flexible duration \cite{lin2018bsn, lin2019bmn, lin2020fast,lin2021learning, bai2020boundary}. To model actions with differing duration, temporal feature pyramid was introduced to generate multi-scale temporal features \cite{lin2017single, zhang2018s3d,liu2019multi, liu2020progressive}. To model dependencies between video segments, different structures have been utilized, including recurrent neural networks \cite{buch2017sst, buch2019end}, graph convolution networks \cite{zeng2019graph, li2020graph, bai2020boundary, xu2020g, zhao2020video}, and transformers \cite{zhang2022actionformer, nawhal2021activity,chang2022augmented}. Distinct from these methods that need exhaustive frame-level annotations, our framework utilizes only point-level annotations. Yet, it effectively captures snippet dependencies and models actions of varying duration.


\textbf{Weakly-supervised TAL.} These methods rely on imprecise or coarse labels during the training stage. They often predict attention scores to pinpoint discriminative action regions and eliminate background frames. Attention scores are typically learned through the Multi-Instance Learning (MIL) scheme \cite{lee2020background, narayan20193c} or via a class-agnostic approach to learn actionness \cite{hong2021cross, luo2021action, narayan2021d2, qu2021acm}. To model the completeness of actions, several methods have proposed complementary learning approaches aimed at discovering different aspects or parts of actions \cite{singh2017hide,zhong2018step, min2020adversarial,zeng2019breaking, liu2019completeness}. Another category of methods relies on an iterative training strategy, which involves generating pseudo-labels from an initial base model to enhance the model's learning capacities \cite{he2022asm,luo2020weakly,pardo2021refineloc, yang2021uncertainty,zhai2020two}. However, these techniques are not capable of generating precise pseudo-labels. Our model generates high-quality pseudo-labels, providing additional guidance to learn the structure of action using slightly more annotations.

\textbf{Point-supervised TAL.} Point-level supervision significantly reduces the cost of annotating action boundaries. Various methods have been proposed to augment annotations: these include the generation of pseudo-actions by expanding annotated frames to their nearby frames \cite{ma2020sf}, or boundary regression based on keyframe prediction \cite{ju2020point}. Other strategies include mining pseudo-background frames from unannotated frames \cite{ma2020sf, lee2021learning} or annotating a random frame from a series of consecutive background frames \cite{yang2021background}. Lee \textit{et al.} \cite{lee2021learning} developed an action-background contrast method for to capture action completeness. CRRC-Net \cite{fu2022compact} proposed a probabilistic pseudo-label mining module to utilize the feature distances from action prototypes to estimate the likelihood of pseudo samples and rectify their corresponding labels for a more reliable classification learning. PCL \cite{li2023prototype} proposed to generate pseudo labels by estimating the semantic similarity of pair-wise frames in the embedding space. FBI-TAL \cite{dong2023fbi} proposed a pseudo-label search strategy by combining foreground and background labels to exploit the information between them and guide the model. 
Li \textit{et al.} \cite{li2024neighbor} uses the relationship of the video segments with their neighbors for pseudo-label generation. Lee \textit{et al.} \cite{lee2023improved} also uses pseudo-labels for action instance boundary learning.

We propose a self-training framework designed to learn the continuous structure of actions with varying duration using our multi-scale transformer, guided by pseudo-labels. Existing work has utilized pseudo-labels to bridge the gap between classification and temporal localization. The advantages of our framework over previous methods are as follows: 1) The simplicity of the pseudo-label generation module, 2) The ability to capture the completeness of actions using self-training, guided by the estimated pseudo-labels, 3) The capability to model actions of varying duration with point-supervision through the design of a feature pyramid, and 4) The integration of a transformer to capture temporal dependencies under limited supervision.


%

\begin{figure*}[t]
\begin{center}
   \includegraphics[width=\textwidth]{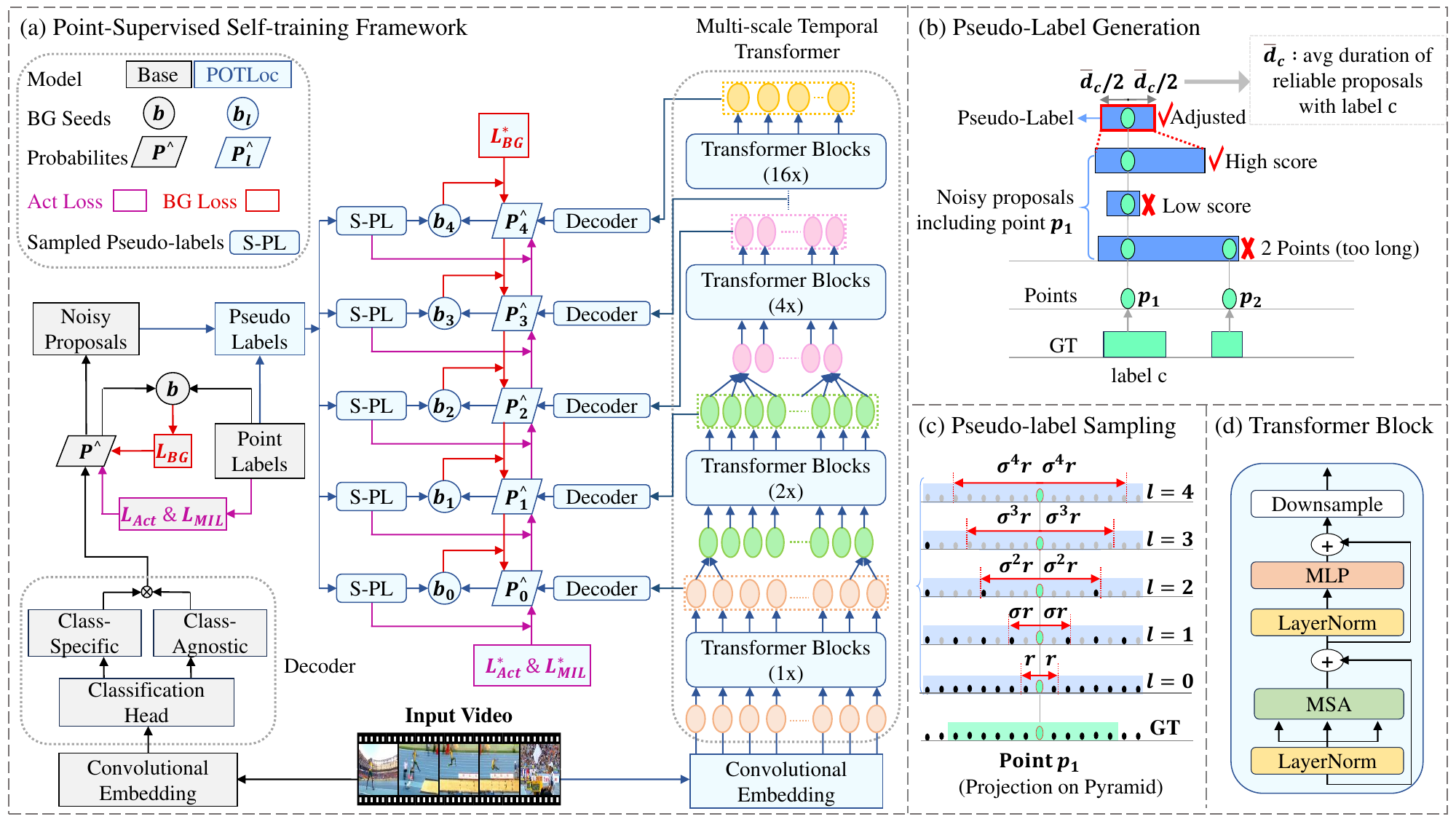}
\end{center}
   \caption{(a) Framework overview. The modules outlined in gray and blue indicate the components of the base and our POTLoc model, respectively. (b) Pseudo-labels are generated from the noisy action proposals predicted by the base model on the training set. The proposals are refined and adjusted based on the point-labels and statistics of the proposals. (c) The pseudo-labels are sampled within a radius around the annotated points at each level $l$ of the pyramid and the block before the pyramid ($l=0$). This sampling helps to mitigate the addition of excessive noise during training, which could be caused by imprecise estimated action boundaries. (a,d) The multi-scale temporal transformer learns to model temporal dependencies and accommodate actions of varying duration when optimized with our enhanced losses, $\mathcal{L}^{\ast}_{\text{MIL}}$, $\mathcal{L}^{\ast}_{\text{Act}}$, and $\mathcal{L}^{\ast}_{\text{BG}}$ supervised with the pseudo-labels.}
\label{fig:framework}
\end{figure*}

\section{Our Proposed Method} 

\textbf{Point-Supervised problem setting}. Given an input video, only a single frame is annotated for each action instance, following \cite{ma2020sf,ju2020point,lee2021learning}. Formally, if there are $N$ action instances in the video, the annotation can be denoted by $\{(\epsilon_i, \Lambda_i) \}_{i=1}^{N}$ where $\epsilon_i$ is the frame index selected from the temporal interval of $i-$th action instance and $\Lambda_i$ is the action label. These annotated time-stamps are referred to as ``points''. Label $\Lambda_i$ is a binary vector where $\Lambda_i[c]$ is equal to 1 if the label of $i-$th action is $c$, and $0$ otherwise. Video-level labels are given by aggregating the labels of annotated points in each video. 

\subsection{Point-Supervised Base Model} 

We employ a base point-supervised model to predict action proposals in the training set. These proposals undergo further refinement, ultimately generating pseudo-labels that serve as augmented supervision for our POTLoc model.





\textbf{Feature extraction and modeling}. The input video is divided into a sequence of snippets, each of which is processed by a pre-trained visual encoder (I3D \cite{carreira2017quo}) for feature extraction. These snippet features are then concatenated to produce a video feature $X$ which is supplied to a shallow temporal convolutional network followed by a sigmoid function. The output results in a class-specific probability signal, $P \in \mathbb{R}^{T \times {C+1}}$, where $p[t,c]$ represents the probability that snippet $t$ belongs to action class $c$. $T$ is the number of video snippets and $C$ is the number of action classes. Additionally, $b_t = p[t,C+1]$ is the probability of background at time $t$. The complement of the background score $b_t$ is the class-agnostic score, denoted by $a_t$. The class-specific and class-agnostic scores are fused to derive the final probability sequence $\hat{P} \in \mathbb{R}^{T \times C+1}$, where $\hat{p}[t,c] = p[t,c] \cdot a_t$ and $\hat{p}[t,C+1] = b_t$.


\textbf{Video-level action prediction}. We predict a video-level probability vector using the class-specific probability sequence $P$. For each action class $c$, we identify the $K$ temporal positions with the highest probability scores. Then, we compute the average score of these positions to represent the video-level probability score for action $c$, denoted as $p^c$. Following the MIL scheme \cite{dietterich1997solving}, a binary cross-entropy loss guides the classification of actions.

\begin{equation}
  \mathcal{L}_{\text{MIL}} = - \sum_{c=1}^{C} \Lambda^c \log(p^c) + (1-\Lambda^c) \log (1-p^c).
\label{eq.MIL}
\end{equation}


\textbf{Snippet-level action prediction}. Given a video with $N$ annotated points, denoted by $\{(\epsilon_i, \Lambda_i) \}_{i=1}^{N}$, a snippet-level focal loss is employed to optimize the probability signal $\hat{P}$ as follows. $\gamma$ is the focusing parameter and is set to $2$. 


\begin{align}
    \footnotesize
    \mathcal{L}_{\text{Act}} = & - \frac{1}{N} \sum_{i=1}^{N} \sum_{c=1}^{C} (1-\hat{p}[\epsilon_i,c])^{\gamma} \Lambda_i[c] \log(\hat{p}[\epsilon_i,c]) \nonumber \\
    & - \hat{p}[\epsilon_i,c]^{\gamma} (1-\Lambda_i[c]) \log (1-\hat{p}[\epsilon_i,c]).
    \normalsize
\label{act-loss}
\end{align}


\textbf{Background modeling}. To differentiate actions from the background, it is crucial to pinpoint the frames that are likely correlated with the background. However, since there are no explicit annotations for these background frames, we implement a method similar to Lee et al. \cite{lee2021learning} to generate ``background seeds" during training. These seeds are chosen from timestamps that have high background scores surpassing a defined threshold. The predicted background seeds in a given video are denoted by $\{t_j \}_{j=1}^{M}$, and $b_{t_j}$ is the probability of background at time $t_j$. At these identified time-steps, we suppress the action probabilities and promote the background probabilities by applying the snippet-level focal loss on signal $\hat{P}$. 

\begin{align}
  \footnotesize
    \mathcal{L}_{\text{BG}} = & - \frac{1}{M} \sum_{j=1}^{M} \Bigl[ 
    \sum_{c=1}^{C} (\hat{p}[t_j,c])^{\gamma} \log (1-\hat{p}[t_j,c]) \nonumber \\
    & + (1-b_{t_j})^{\gamma} \log b_{t_j}].
    \normalsize
\label{bg-loss}
\end{align}



%





\textbf{Joint training}. The total loss for the base model is a weighted combination of the three aforementioned losses, calculated as follows, where $\lambda_{\star}$ terms balance the losses and are determined through empirical analysis.

\begin{align}
  L_{\text{Total}} = \lambda_{\text{MIL}}\mathcal{L}_{\text{MIL}} + \lambda_{\text{Act}} \mathcal{L}_{\text{Act}} + \lambda_{\text{BG}} \mathcal{L}_{\text{BG}}.  
\end{align}

\textbf{Action proposal generation}. We set a threshold on the predicted video-level scores to identify the action categories present in the video. Then, we apply a threshold on the snippet-level action scores $\hat{P}$ for those action categories already predicted. We then merge consecutive candidate segments to form proposals, each of which is assigned a confidence score based on its outer-inner-contrast score \cite{lee2020background}. Finally, we use the non-maximum suppression (NMS) technique to eliminate overlapping proposals.

\subsection{Pseudo-label Generation from Action Proposals}

The base model utilizes point-supervision to predict initial action proposals on the training set. These proposals are redundant and noisy and are unsuitable as pseudo-labels for self-training. In this section, we propose an algorithm to generate pseudo-labels by leveraging the statistics of the proposals and their locations in relation to the annotated points. 




\textbf{Proposal formulation}. We define $S$ to be the set of predicted proposals on the training set $\mathcal{V}$. For each video $v \in \mathcal{V}$, $S_v$ denotes the predicted proposal and $\varrho_v$ denotes the points. The predicted start, end, label, and confidence score of the $j$-th proposal $\varphi_j$ are denoted by $s_j$, $e_j$, $\Lambda_j$, and ${cs}_j$, respectively. Also, $p_i$ is the $i$-th point with time $\epsilon_i$ and label $\Lambda_i$.

\begin{equation}
 S_v = \cup_{j} \{\varphi_j= (s_j, e_j, \Lambda_j, {cs}_j)\} \ , \ \varrho_v = \cup_{i} \{p_i = (\epsilon_i, \Lambda_i)\}.  
\end{equation}

\textbf{Pseudo-label formulation}. For a video $v$, the pseudo-label set is defined as $S^\ast_v = \{ (\epsilon_n, s_n,e_n,\Lambda_n) \}_{n=1}^{N_v} $ where $N_v$ is the number of annotated points in the video and $\epsilon_n$ denotes the $n$-th point. The estimated start, end, and label of the $n$-th pseudo-label are denoted by $s_n$, $e_n$, and $\Lambda_n$, respectively. One pseudo-label is generated for each annotated point $\epsilon_n$ such that $\epsilon_n \in [s_n,e_n]$. Set $S^{\ast}$ is the union of pseudo-labels for all training videos. 

\begin{equation}
 S^{\ast} = \cup_{n}\{(\epsilon_n, s_n,e_n,\Lambda_n) \}.
 \label{s*-def}
\end{equation}

\begin{algorithm}[t!]
 \caption{\textit{Pseudo-label Generation}}
 \label{algo:refine}
  \textbf{Input:} Proposals $S_v$, points $\varrho_v$, for each $v \in \mathcal{V}$ \\
  $S_v = \bigcup_{j} \{\varphi_j= (s_j, e_j, \Lambda_j, {cs}_j)\}$, $\varrho_v = \bigcup_{i} \{p_i = (\epsilon_i, \Lambda_i)\}$ \\
\textbf{Output:} Pseudo-labels $S^{\ast} = \bigcup_{n}\{(\epsilon_n, s_n,e_n,\Lambda_n) \}$ \\ 
\textit{Initialization:} $S^{\ast} = \emptyset$
 \begin{algorithmic}[1]
\Function{f}{$ \varphi_j, p_i$}  \Comment (Check if $p_i$ belongs to $\varphi_j$)
\If {$(\Lambda_i = \Lambda_j) \cap (s_j \leq \epsilon_i \leq e_j)$} \textbf{return} \textit{True}
\Else \textbf{ return} \textit{False}
\EndIf
\EndFunction
 \For {$\varphi_j \in S_v$ and $v \in \mathcal{V}$:}
        \If {$|\{p_i | p_i \in \varrho_{v} \text{ s.t. } F(\varphi_j, p_i)  \}| = 1$} 
        \State $S^{\ast} = S^{\ast} \cup \{(\epsilon_i, s_j,e_j,\Lambda_j, cs_j) \}$
        \EndIf
 \EndFor
 \For {$c = 1$ to $C$:}
   \State  $\Bar{d}_c = \text{mean}(\{(e_j-s_j) | \varphi_j \in S^{\ast} \text{ s.t. } (\Lambda_j[c]=1) \})$
 \EndFor 
 \For {$p_i$ in $\varrho_v$ and $v \in \mathcal{V}$:}
        \State $\Delta = \Bar{d}_{c}/2$ \text{ s.t. } $\Lambda_i[c]=1$
        \If {$\{ \varphi \in S^{\ast} | F(\varphi, p_i)  \} = \emptyset$}
            \State $\tau = \{ \varphi \in S | F(\varphi, p_i)\} $
            \State $k = \text{Argmax}_{cs} (\tau) $  \Comment (proposal with max score)
            \State $s_k = \max(s_k, \epsilon_i -\Delta)$ , $e_k = \min(e_k, \epsilon_i + \Delta)$ 
            \State $S^{\ast} = S^{\ast} \cup \{(\epsilon_i, s_k,e_k,\Lambda_k, cs_k)\}$
        \Else
            \State $\tau = \{ \varphi \in S^{\ast} | F(\varphi, p_i)\} $
            \State $k = \text{Argmax}_{cs} (\tau) $  \Comment (proposal with max score)
            \State $S^{\ast} = (S^{\ast} - \tau) \cup \{(\epsilon_i, s_k,e_k,\Lambda_k, cs_k)\}$
        \EndIf
 \EndFor
  \end{algorithmic} 
 \end{algorithm}

\textbf{Pseudo-label generation algorithm}. Initially, set $S^{\ast}$ only includes the proposals from $S$ that contain precisely one annotated point. These selected proposals are considered more reliable because they are neither excessively long nor too short, avoiding multiple or no points at all. Within $S^{\ast}$, for each action class $c$, the average duration of proposals with label $c$, denoted as $\Bar{d}_c$, is calculated. We must ensure that each annotated point exists in at least one proposal for complete coverage. Suppose there is a point belonging to action class $c$ with timestamp $\epsilon_i$ that is not included in any of the proposals in set $S^{\ast}$. In this case, we search for a list of proposals with label $c$ in the initial set $S$ that include point $\epsilon_i$ and select the one with the highest confidence score. We truncate this proposal within a distance of $\Bar{d}_c/2$ from $\epsilon_i$ to prevent the new proposal from being too long. All the newly generated proposals from this step are added to set $S^{\ast}$. Finally, we ensure that each annotated point belongs to exactly one proposal by keeping the proposal with the highest confidence score that contains the point and removing the rest. The confidence scores ${cs}_n$ were only used for pseudo-label generation and are discarded from $S^{\ast}$ at the end. The details of this procedure is summarized in \textit{Algorithm \ref{algo:refine}}.



\subsection{Pseudo-label Oriented Multi-scale Transformer}

Fig.~\ref{fig:framework}(a) provides an overview of our framework. The base model utilizes point-supervision to predict initial action proposals. These preliminary action proposals are subsequently used to generate pseudo-labels for the training set based on the point-labels and the statistics of the proposals, as shown in Fig.~\ref{fig:framework}(b). Our POTLoc model employs a multi-scale temporal transformer to capture the temporal dependencies within video snippets and to learn multi-scale temporal action instances, Fig.~\ref{fig:framework}(a,d). POTLoc, supervised by the pseudo-labels, is optimized with three enhanced loss functions, $\mathcal{L}^{\ast}_{\text{MIL}}$, $\mathcal{L}^{\ast}_{\text{Act}}$, and $\mathcal{L}^{\ast}_{\text{BG}}$, to separate actions from background and discriminate actions. Since the pseudo-labels are imprecise estimations of the action boundaries, we sample from the pseudo-labels to mitigate the potential addition of excessive noise during training, Fig.~\ref{fig:framework}(c). The pseudo-labels play a crucial role by equipping the network with detailed information about the approximate locations of actions. This process enhances the effective use of the transformer model and feature pyramid, thereby improving the model's ability to understand action dynamics.

\textbf{Multi-scale temporal transformer.} Given an input video, we extract snippet-level visual features with a pre-trained visual encoder and concatenate them to generate a video feature $X \in \mathbb{R}^T$, where $T$ is the number of snippets. Each snippet feature is embedded using a shallow temporal convolutional network with layer normalization and ReLU, resulting in feature vector $Z^0 \in \mathbb{R}^{T \times D}$. This feature is the input to the transformer network which is employed to model the temporal dependencies between snippets. Feature $Z^0$ is projected using learnable parameters $W_Q \in \mathbb{R}^{D \times D_q}, W_K \in \mathbb{R}^{D \times D_k},$ and $W_V \in \mathbb{R}^{D \times D_v}$ to extract query, key, and values features, denoted by $Q, K,$ and $V$, respectively, with $D_q=D_k$. The output of self-attention is $S = \text{Softmax} (QK^T/\sqrt{D_q}) V$ where $S \in \mathbb{R}^{T \times D}$. We adapt the local self-attention within a window to reduce the time and memory complexity, following \cite{zhang2022actionformer}. The transformer network consists of several layers, wherein each layer is composed of multiheaded self-attention (MSA) and MLP blocks, with GELU activation. To model multi-scale features for actions with different duration, we implement down-sampling between transformer blocks using a strided depthwise 1D convolution, resulting in a temporal feature pyramid $Z =\{Z^1,Z^2,\cdots,Z^L\}$. The $l$-th transformer block receives the input feature $Z^{l-1}$ and returns the feature $Z^l$, where $Z^l \in \mathbb{R}^{T_l \times D}$, $T_l = T/{\sigma^l}$, and $\sigma$ is the down-sampling ratio. The input to the first transformer block is $Z^0$. Feature pyramid captures multi-scale temporal information, enabling the model to capture both short-term and long-term temporal dependencies, leading to a more comprehensive representation of action dynamics.

\textbf{Action decoder}. A shallow 1D convolutional network with layer normalization and ReLU is attached to each pyramid level with its parameters shared across all levels. A sigmoid function is attached to each output dimension to predict the probability of actions and background. The output of the $l$-th level of the feature pyramid is a probability sequence, denoted by $P_l \in \mathbb{R}^{ T_{l}\times {C+1}}$, where $T_l$ is the temporal dimension on the $l$-th level. Furthermore, $b_{l,t} = p_l[t,C+1]$ is the probability of background at time $t$ on level $l$. The class-specific scores are fused with the class-agnostic scores to derive the final probability sequence $\hat{P_l} \in \mathbb{R}^{T_l \times C+1}$. 




\textbf{Pseudo-label sampling}. We only consider a narrow interval around the annotated point $\epsilon_n$ as a positive instance, as shown in Fig.~\ref{fig:framework}(c). The interval $[\epsilon_n-r, \epsilon_n+r]$ is sampled from pseudo-label interval $[s_n,e_n]$, where $r$ represents the sampling radius and is selected empirically. This sampling procedure mitigates the potential addition of excessive noise during the training, as the interval $[s_n,e_n]$ is merely an approximation of the action boundaries. The projection of the pseudo-label $(\epsilon_n, s_n,e_n,\Lambda_n)$ onto the $l$-th level of the pyramid becomes $(\epsilon_n/\sigma^l,s_n/\sigma^l,e_n/\sigma^l,\Lambda_n)$ where $\sigma$ represents the down-sampling ratio. For each level $l$, we sample an interval with radius $\sigma^l \cdot r$ that is centered around the projected point and located within the projected boundaries. The  pseudo-labels at level $l$ are denoted as following. 

\begin{equation}
    S^{\ast}_l = \cup_{n}\{(\epsilon_n^l, s_n^l,e_n^l,\Lambda_n) \}.
\end{equation}

\textbf{Video-level action prediction}. The video-level score for class $c$ is defined as the average of $p_l[t^c_{l,k},c]$ scores where $\{t^c_{l,k}\}_{k=1}^{K}$ are the top-$K$ positions on level $l$. The average is calculated over all levels of the pyramid. We utilize the MIL loss (eq.~\ref{eq.MIL}) for this extended version and name it $\mathcal{L}^{\ast}_{\text{MIL}}$. 



\textbf{Snippet-level action prediction}. To simplify the notations, for each level $l$, we collect all temporal positions of all pseudo-labels into a set, denoted by $\Phi^l$, as follows.

\begin{equation}
    \Phi^l = \mathop{\cup}_n \{(t,\Lambda_n) | \ t \in  [s^l_n,e^l_n] \text{ for }  (\epsilon_n^l, s^l_n,e^l_n,\Lambda_n) \in S^{\ast}_l\}.
\end{equation}

We subsequently rename the elements in $\Phi^l$ as $\Phi^l = \{(t_m, \Lambda_m)\}_{m=1}^{M_l}$. We extend the snippet-level focal loss (Eq.~\ref{act-loss}) to all temporal positions of all pseudo-labels to optimize the learning of the probability signal $\hat{P_l}$ for each level $l$ of the pyramid. $M$ is the total number of positive instances.

\begin{equation}
\begin{aligned}
\mathcal{L}^*_{\text{Act}} = -\frac{1}{M} &\sum_{l=1}^{L} \sum_{m=1}^{M_l} \sum_{c=1}^{C} (1-\hat{p}_l[t_m,c])^{\gamma} \Lambda_m[c] \log(\hat{p}_l[t_m,c]) \\
&- \hat{p}_l[t_m,c]^{\gamma} (1-\Lambda_m[c]) \log (1-\hat{p}_l[t_m,c]).
\end{aligned}
\label{losses*}
\end{equation}


\textbf{Background modeling}. To distinguish actions from the background, similar to the base model, we select the temporal positions not belonging to any of the pseudo-labels and possessing a background probability exceeding a certain threshold on each level $l$ of the pyramid. The background loss presented in Eq.~\ref{bg-loss} is extended to all pyramid levels to optimize the probability signal $\hat{P_l}$, and is denoted by $\mathcal{L}^{\ast}_{\text{BG}}$.\\

\textbf{Joint training}. Our POTLoc model is trained using a combination of the three enhanced losses with $\lambda_{\star}$ weighting parameters that are determined through empirical analysis.

\begin{equation}
\begin{aligned}
L_{\text{Total}} = \lambda_{\text{MIL}}\mathcal{L}^{\ast}_{\text{MIL}} + \lambda_{\text{Act}} \mathcal{L}^{\ast}_{\text{Act}} + \lambda_{\text{BG}} \mathcal{L}^{\ast}_{\text{BG}}.
\end{aligned}
\end{equation}



\textbf{Inference}. The action categories are identified using the video-level scores. The action proposals are predicted from all pyramid levels by applying thresholds to the snippet-level action scores $\hat{P_l}$ for each level $l$ for the predicted classes. The strategy used is similar to the inference of the base model.



\section{Experiments}

\subsection{Experimental Setting}

\textbf{Datasets}. THUMOS14 consists of untrimmed videos spanning 20 distinct categories. Following previous methods \cite{lee2021learning,he2022asm}, we utilized the validation set for training, and the testing set for evaluation. ActivityNet-v1.2 is a large-scale dataset encompassing 100 complex daily activities. We follow the convention of using the training set to train our model, and the validation set for evaluation \cite{lee2021learning,he2022asm}.

\textbf{Evaluation metric}. The Mean Average Precision (mAP) under different Intersection over Union (IoU) thresholds is utilized as the evaluation metric, wherein the Average Precision (AP) is computed for each action class. 


\textbf{Implementation details}. For feature extraction, the two-stream I3D model \cite{carreira2017quo} is utilized on both datasets. Segments consisting of $16$ consecutive frames are fed as input to the visual encoder, employing a sliding window approach with a stride of $16$ on both THUMOS14 and ActivityNet-v1.2. Both base and POTLoc models are optimized by Adam  \cite{kingma2014adam} with the learning rate of $10^{-4}$ for $50$ epochs. In the base model, the original number of feature segments is used without sampling. However, in the main model, the input length is set to $768$ for THUMOS14 and to $192$ for ActivityNet-v1.2, using random sampling and linear interpolation. A window of $12$ and $7$ is used for local self-attention on THUMOS14 and ActivityNet-v1.2, respectively. In POTLoc model, the number of pyramid levels is set to $l=2$ and the sampling radius is set to $r=2$. The parameter $r$ is defined on the feature grid, representing the distance in terms of the number of features. The batch size is set to $4$ on THUMOS14, and to $64$ on ActivityNet-v1.2. At inference time, the full sequence is fed into the model without sampling. 


\textbf{Computational Complexity:} The number of parameters and FLOPs are $17.9$B and $12.6$M for the base model and $37.5$B and $24.4$M for multi-scale transformer.


\subsection{Comparison with State-of-the-art Methods}

Table \ref{SOTA-thumos} presents a comprehensive comparison with the state-of-the-art methods on THUMOS'14 and ActivityNet-v1.2.

\textbf{Results on THUMOS’14}. 
Our model achieves state-of-the-art performance among weakly-supervised and point-supervised methods in terms of average mAP. Additionally, our model demonstrates remarkable results of an $6\%$ average mAP increase compared to weakly-supervised methods, despite only using slightly more annotations. 



\textbf{Results on ActivityNet-v1.2}. Our model outperforms all the state-of-the-art weakly and point-supervised methods in terms of mAP, consistently across all the IoU thresholds. We note that the performance gains over weakly-supervised methods on ActivityNet-v1.2 are smaller compared to those on the THUMOS'14 dataset. This is primarily because the average number of action instances per video in THUMOS'14 is higher than that in ActivityNet-v1.2 (15.5 vs. 1.5). Consequently, on THUMOS'14, the model can learn to distinguish actions from the background with the assistance of inferred background seeds situated between consecutive action points. This is more challenging on ActivityNet-v1.2 due to the sparse nature of action instances.

\begin{table*}[t!]
\centering
\resizebox{\columnwidth}{!}{%
\begin{tabular}{c l  c c c c c  c | c c c c}
 \hline
 \multicolumn{1}{c}{\multirow{3}{*}{Group}} &
\multicolumn{1}{l}{\multirow{3}{*}{Method}} &
 \multicolumn{6}{c}{THUMOS'14} &
  \multicolumn{4}{|c}{ActivityNet-v1.2} \\ \cline{3-8} \cline{9-12}

\multicolumn{1}{c}{}  &  \multicolumn{1}{c}{} &  \multicolumn{5}{c}{mAP@IoU (\%)}& mAP-AVG& 
\multicolumn{3}{c}{mAP@IoU (\%)} & mAP-AVG\\ \cline{3-8} \cline{9-12}

&   &  0.3	&0.4	&0.5	&0.6	&0.7 & (0.1:0.7) & 0.5 & 0.75 & 0.95 & (0.5:0.95) \\
\hline

 







%



 \multicolumn{1}{c}{\multirow{15}{*}{WS}} & ASL\cite{ma2021weakly} &51.8  & - & 31.1 & - & 11.4 & 40.3 & 40.2 & - & - & 25.8 \\

&CoLA \cite{zhang2021cola} & 51.5 & 41.9 & 32.2 & 22.0 & 13.1  & 40.9 &   42.7 &  25.7 &  5.8 & 26.1 \\


& AUMN \cite{luo2021action} & 54.9 &  44.4 & 33.3 &20.5  &9.0   & 41.5 & 42.0 &  25.0 & 5.6 &  25.5 \\


& FTCL \cite{gao2022fine} & 55.2 & 45.2 & 35.6 & 23.7 & 12.2  & 43.6  & - & - & - & -\\

& UGCT \cite{yang2021uncertainty}  &  55.5 & 46.5 & 35.9 & 23.8 & 11.4  & 43.6 &  41.8 &  25.3 &  5.9 &  25.8  \\ 

&CO2-Net \cite{hong2021cross}& 58.2 & 47.1 & 35.9 & 23.0 & 12.8 & - & 43.3 &  26.3 &  5.2 & 26.4 \\

 &  D2-Net \cite{narayan2021d2} &   52.3  &  43.4  & 36.0 & - & -   & - & 42.3 &  25.5 &  5.8 &  26.0 \\

& ASM-Loc\cite{he2022asm}   &   57.1 & 46.8 & 36.6 & 25.2 & 13.4  &   45.1 & - & - & - & - \\ 

& RSKP\cite{huang2022weakly}  & 55.8 & 47.5 & 38.2 & 25.4 & 12.5 &   45.1& - & - & - & -\\

 & TS\cite{wang2023two}  & 60.0 & 47.9 & 37.1 & 24.4 & 12.7 & 46.2& - & - & - & -\\

& DELU\cite{chen2022dual} & 56.5 & 47.7 & 40.5 & 27.2 & 15.3  &  46.4 & 44.2 & 26.7 & 5.4 & 26.9 \\

& P-MIL \cite{ren2023proposal} & 58.9 & 49.0 & 40.0 & 27.1 & 15.1  &  47.0 &  44.2 & 26.1 & 5.3 & 26.5 \\

& Zhou \textit{et al.} \cite{zhou2023improving} & 60.7 & 51.8 & 42.7 & 26.2 & 13.1 & 48.3 & - & - & - & - \\

& PivoTAL \cite{rizve2023pivotal} & 61.7 & 52.1 & 42.8 & 30.6 & 16.7  & 49.6 & - & - & - & -\\  

\hline
\hline

\multicolumn{1}{c}{\multirow{9}{*}{PS}} & SF-Net \cite{ma2020sf} &   52.8 & 42.2 & 30.5 & 20.6 & 12.0  & 41.2 & 37.8 & - & - & 22.8 \\ 

& DCM \cite{ju2021divide}  & 58.1 & 46.4 & 34.5 & 21.8 & 11.9 & 44.3& - & -& -& -\\

& PTAL \cite{ju2020point}  &  58.2 & 47.1 & 35.9 & 23.0 & 12.8 & - & -& -& -& -\\ 

& BackTAL \cite{yang2021background}  &  54.4 & 45.5 & 36.3 & 26.2 & 14.8 &  - & 41.5 & 27.3  & 4.7 & 27.0\\

& PCL \cite{li2023prototype} & {63.3}  & {55.9} &  {44.4} & - & - & - & - & -& - & - \\

& Lee \textit{et al.} \cite{lee2021learning}  & 64.6 & 56.5 & 45.3 & 34.5 & 21.8  & 52.8 & 44.0 & 26.0 & 5.9  & 26.8 \\ 

& CRRC-Net \cite{fu2022compact} &  {67.1} & {57.9} & {46.6} & {33.7} & {19.8} & {53.8} & - & - & - & - \\

& Lee \textit{et al.} \cite{lee2023improved} & {66.8}  & {57.8} &  {47.1} & {34.8} & {21.1} & {-} & {44.6} & {26.7} & {6.1}  & {27.2} \\

& FBI-TAL \cite{dong2023fbi} & {66.7}  & {58.3} &  {48.3} & {36.3} & {21.9} & {54.6} & - & - & - & - \\

& Li \textit{et al.} \cite{li2024neighbor} & {66.6}  & {59.4} &  {48.6} & {36.7} & \textbf{22.7} & {55.1} & {43.4} & {31.3} & {5.4} & {27.5} \\


& \textbf{POTLoc} & {\textbf{68.8}} & {\textbf{59.5}} & {\textbf{50.1}} & {\textbf{37.1}} & 21.2 & {\textbf{55.7}} & \textbf{45.1} & \textbf{27.6}& \textbf{6.8} & \textbf{28.0}\\

\hline 
\end{tabular}}
\caption{Comparison with weakly-supervised (WS) and point-supervised (PS) methods on THUMOS'14 and ActivityNet-v1.2. The results are reported in terms of mAP (\%) at different tIoU thresholds. The bold numbers show the best results. }
\label{SOTA-thumos}
\end{table*}





\subsection{Ablation Studies}
We conduct ablation studies on THUMOS'14 to analyze the impact of each component of the proposed model. 

\textbf{Pseudo-label generation}. Table \ref{PL-generation-table} demonstrates the quality of the generated pseudo-labels. This table reports the performance on the train set (validation set) of THUMOS'14. $\alpha$ represents the ratio of the number of generated proposals to the ground-truth instances. The noisy proposals are predicted by the base model without refinement. The refinement significantly removes redundant proposals and improves the alignments with ground-truth intervals. The pseudo-labels provide exactly one interval around each annotated point and $\alpha=1$. Furthermore, the performance of the pseudo-labels is $6\%$ average mAP higher than the noisy proposals which highlights the effectiveness of the proposed pseudo-label generation method.

\begin{table}[t!]
\centering
\resizebox{0.8\columnwidth}{!}{%
\begin{tabular}{c | c | c | c  }
 \hline
\multicolumn{1}{c|}{\multirow{2}{*}{Step}} & \multicolumn{1}{c|}{\multirow{2}{*}{$\alpha= \frac{\text{\#Proposals}}{\text{\#GT}}$}} & 
\multicolumn{1}{c|}{\multirow{2}{*}{mAP@0.5 (\%)}} & 
mAP-AVG (\%)\\ 
& &  & (0.1:0.7) \\
\hline
Noisy Proposals & $\sim$ 12  & 57.0 &   63.5 \\ 
Pseudo-labels & $\sim$ 1 & 62.7 &  69.5 \\
\hline 
\end{tabular}}
\caption{Analysis of pseudo-labels on the \textbf{validation set} of THUMOS'14. }
\label{PL-generation-table}
\end{table}

\begin{table*}[t!]
\centering
\resizebox{0.8\hsize}{!}{
\begin{tabular}{ c | c | c | c  }
\hline
Backbone, Losses & Supervision & SL   & mAP(\%)  \\
\hline
\multicolumn{1}{c|}{\multirow{7}{*}{\parbox{4.5cm}{ Multi-scale Transformer, \\ Enhanced Losses: \\ $\{\mathcal{L}^{\ast}_{\text{MIL}}, \mathcal{L}^{\ast}_{\text{Act}}, \mathcal{L}^{\ast}_{\text{BG}}\}$}}} & \multirow{2}{*}{Ground-truth} & $\checkmark$ &  56.0 \\ 

& & \ding{55} &   52.1 \\\cline{2-4}

&  \multirow{2}{*}{\parbox{2.5cm}{Pseudo-labels \\  \textbf{(POTLoc)}}} & $\checkmark$ &  \textbf{55.7}  \\

&  & \ding{55} &  51.0  \\\cline{2-4}

&  \multirow{2}{*}{Noisy Proposals} & $\checkmark$ & 26.8  \\

&  & \ding{55} &  38.3 \\\cline{2-4}

&  Points & \ding{55}  & 50.4 \\

\hline
\hline

\multicolumn{1}{c|}{\multirow{7}{*}{\parbox{4.5cm}{Temporal Convolutions, \\ Base Losses: \\ $\{\mathcal{L}_{\text{MIL}}, \mathcal{L}_{\text{Act}}, \mathcal{L}_{\text{BG}}\}$}}} &
\multirow{2}{*}{Ground-truth}   & $\checkmark$ & 50.1  \\
&  & \ding{55} &  44.8 \\\cline{2-4}
&  \multirow{2}{*}{Pseudo-labels} & $\checkmark$ &  49.8  \\
 &  & \ding{55} &   46.9 \\\cline{2-4}
&  \multirow{2}{*}{Noisy Proposals} & $\checkmark$ &   28.7  \\
&  & \ding{55} &   38.2 \\\cline{2-4}
&  Points & \ding{55} &  47.4 \\
\hline
\end{tabular}}
\caption{Impact of the main components of our framework on THUMOS'14, measured in terms of average mAP. SL denotes sampling, with the radius set to 2. The bold number represents the performance of our full POTLoc model.} 
\label{ablation-main}
\end{table*}

\textbf{Impact of pseudo-labels}. Table \ref{ablation-main} highlights the crucial role of pseudo-label generation. It is worth noting that the use of noisy proposals results in poor performance, underperforming the models supervised with single points. This is because noisy proposals provide a highly inaccurate estimation of action boundaries. Moreover, many of these proposals may be redundant and overlapping. This highlights the importance of proposal refinement in our pseudo-label generation. To further assess the quality of the pseudo-labels, we conduct experiments using ground-truth labels. We observe that the performance of the model supervised by pseudo-labels is comparable with that of the fully-supervised model. This can be attributed to our model not depending on information about the precise location of action boundaries, which could otherwise be employed in a regression loss.

\textbf{Label sampling}. The impact of sampling across different supervision levels is demonstrated in Table \ref{ablation-main}. Sampling consistently improves the performance for both pseudo-labels and ground-truth labels. When pseudo-labels are utilized, sampling mitigates the noise introduced by imprecise action boundaries. Moreover, when using ground-truth labels, sampling encourages higher scores around action centers, encouraging the model to learn meaningful and representative action snippets. In other words, sampling selects snippets that are closer to the action centers (often more indicative of the action) while avoiding boundary snippets that can be ambiguous or contain transitional movements not representative of the action. Therefore, sampling improves the performance even in the case of training with ground-truth boundaries. However, sampling does not enhance performance when the model is supervised with noisy proposals. This is primarily because the center of the noisy proposals may not necessarily be close to the center of the action instances. In this scenario, sampling may inadvertently lead to a focus on a random video snippet such as background. Moreover, Table \ref{radius-table} demonstrates the importance of the sampling strategy with different sampling radius $r$, which $r= \infty$ indicates \textit{no sampling}.


\begin{table}[t!]
\centering
\resizebox{0.8\columnwidth}{!}{%
\begin{tabular}{c | c c c c c | c }
 \hline
 \multicolumn{1}{c|}{\multirow{2}{*}{Radius}}  &  \multicolumn{5}{c|}{mAP@IoU (\%)} & mAP-AVG (\%)\\ \cline{2-7}
  &  0.3	&0.4	&0.5	&0.6	&0.7 & (0.1:0.7) \\
\hline
$r=2$ & \textbf{68.8} & \textbf{59.5} & \textbf{50.1} & \textbf{37.1} & \textbf{21.2} & \textbf{55.7} \\
$r=4$ & 65.4 & 56.8 & 46.0 & 34.0 & 18.7 & 53.0 \\
$r= \infty$ & 63.0 & 53.8 & 43.5 & 32.3 & 18.4 & 51.0\\
\hline 
\end{tabular}}
\caption{Impact of the pseudo-label sampling radius $r$ in POTLoc model on THUMOS'14.}
\label{radius-table}
\end{table}

\textbf{The backbone architecture}. Table \ref{ablation-main} illustrates that the multi-scale transformer when trained with enhanced losses, achieves significantly better results compared to the base model. The latter only consists of convolutional layers and is trained with base losses. The performance enhancement is consistent across ground-truth, pseudo-labels, and points supervision. However, for noisy proposals, the results of different models are comparable. Table \ref{ablation-fpn-level} demonstrates the impact of the number of pyramid levels, denoted by $l$, in POTLoc. The model denoted by $l=0$ incorporates transformer blocks without a feature pyramid, leading to the lowest performance. The model with $l=2$ achieves the highest performance at an IoU of $0.7$ reflecting generation of complete action proposals with the assistance of the feature pyramid. Our findings suggest that adding more pyramid levels ($l\geq 3$) does not improve the performance further. 

\begin{table}[t!]
\centering
\resizebox{0.7\hsize}{!}{
\begin{tabular}{ c | c c c c c c }
\hline
Levels & $l=0$ & $l=1$ & $l=2$ & $l=3$ & $l=4$ \\
\hline
mAP@0.7 & 18.8 & 18.4 & \textbf{21.2} &  20.7 & 19.25 \\
\hline
mAP-AVG & 51.2 & 54.7 & \textbf{55.7} & 55.7 & 54.1 \\
\hline
\end{tabular}}
\caption{Impact of the number of pyramid levels (denoted by $l$) on THUMOS'14. The backbone is POTLOC’s multi-scale transformer supervised with pseudo-labels.}
\label{ablation-fpn-level}
\end{table}

\textbf{Impact of the loss functions}. 
As mentioned earlier, our POTLoc model is trained using a combination of three loss functions $\mathcal{L}^{\ast}_{\text{MIL}}$, $\mathcal{L}^{\ast}_{\text{Act}}$, and $\mathcal{L}^{\ast}_{\text{BG}}$ (eq.~\ref{losses*}). Table \ref{ablation-losses} reports the impact of the $\lambda_{\star}$ weighting parameters. The highest average mAP is achieved when $\lambda_{\text{MIL}}=1$, $\lambda_{\text{Act}}=0.5$ and $\lambda_{\text{BG}}=1$.


\begin{table}[t!]
\centering
\resizebox{0.5\hsize}{!}{
\begin{tabular}{c c c | c}
 \hline

\multicolumn{3}{c|}{Loss Parameters} & mAP-AVG (\%) \\

\hline
$\lambda_{\text{MIL}}$ & $\lambda_{\text{Act}}$ & $\lambda_{\text{BG}}$ & (0.1:0.7)\\
\hline
$1$ & $1$ & $1$ & $55.2$ \\ 
$0.5$ & $1$ & $1$ & $54.2$  \\ 
$1$ & $0.5$ & $1$ & \textbf{55.7}  \\
$1$ & $1$ & $0.5$ &  $52.7$  \\ 
\hline
\end{tabular}}
\caption{Impact of the loss functions in POTLoc on THUMOS'14.}
\label{ablation-losses}
\end{table}


\textbf{Distribution of annotated points}. In the point-supervision setting, only a single frame per action instance is annotated in the training set. SF-Net \cite{ma2020sf} proposed to simulate point annotations by sampling a single frame for each action instance. The Uniform distribution method randomly selects a frame within the action boundaries of each action, while the Gaussian distribution method does so with respect to a given mean and standard deviation. Typically, the Gaussian distribution is more likely to sample frames closer to the central timestamps of actions, thereby increasing the chances of choosing a more discriminative snippet. In contrast, the Uniform distribution can sample frames from any part of the action, without this central bias. Table \ref{point-distribution} demonstrates that POTLoc attains state-of-the-art results with both Uniform and Gaussian point-level distributions on THUMOS'14, indicating its robustness. However, it is observed that the POTLoc's performance is lower with the Uniform distribution as compared to the Gaussian distribution. We conjecture this may be attributed to the Uniform distribution's tendency to select less discriminative snippets for point annotation, which can occur anywhere within the action's extent, such as at the boundaries. Bridging the performance gap between models trained with different sampling distributions of annotated points (Gaussian and Uniform) can be considered for future work.

\begin{table}[t!]
\centering
\resizebox{0.8\columnwidth}{!}{%
\begin{tabular}{c | c | c  c c | c }
 \hline
 \multicolumn{1}{c|}{\multirow{2}{*}{Distribution}}  &  \multicolumn{1}{c|}{\multirow{2}{*}{Method}} & \multicolumn{3}{c|}{mAP@IoU (\%)} & mAP-AVG (\%)\\ \cline{3-6}
  & &  0.3	&0.5	&0.7 & (0.1:0.7) \\
\hline
\multicolumn{1}{c|}{\multirow{2}{*}{Gaussian}} & \textbf{POTLoc} & \textbf{68.8} &  \textbf{50.1}  & \textbf{21.2} & \textbf{55.7} \\
 & LACP \cite{lee2021learning} & 64.6  & 45.3 & \textbf{21.8} & 52.8 \\    
\hline
\multicolumn{1}{c|}{\multirow{2}{*}{Uniform}} & 
\textbf{POTLoc} & \textbf{64.1} &  \textbf{43.5}  & 17.7 & \textbf{51.3} \\  
 & LACP \cite{lee2021learning} & 60.4 & 42.6 &  \textbf{20.2} & 49.3 \\ 
\hline 
\end{tabular}}
\caption{Performance comparison with uniform and Gaussian point-level distributions on THUMOS'14. }
\label{point-distribution}
\end{table}

\begin{figure}[t!]
    \centering
    \begin{subfigure}[b]{0.8\linewidth}
        \includegraphics[width=\linewidth]{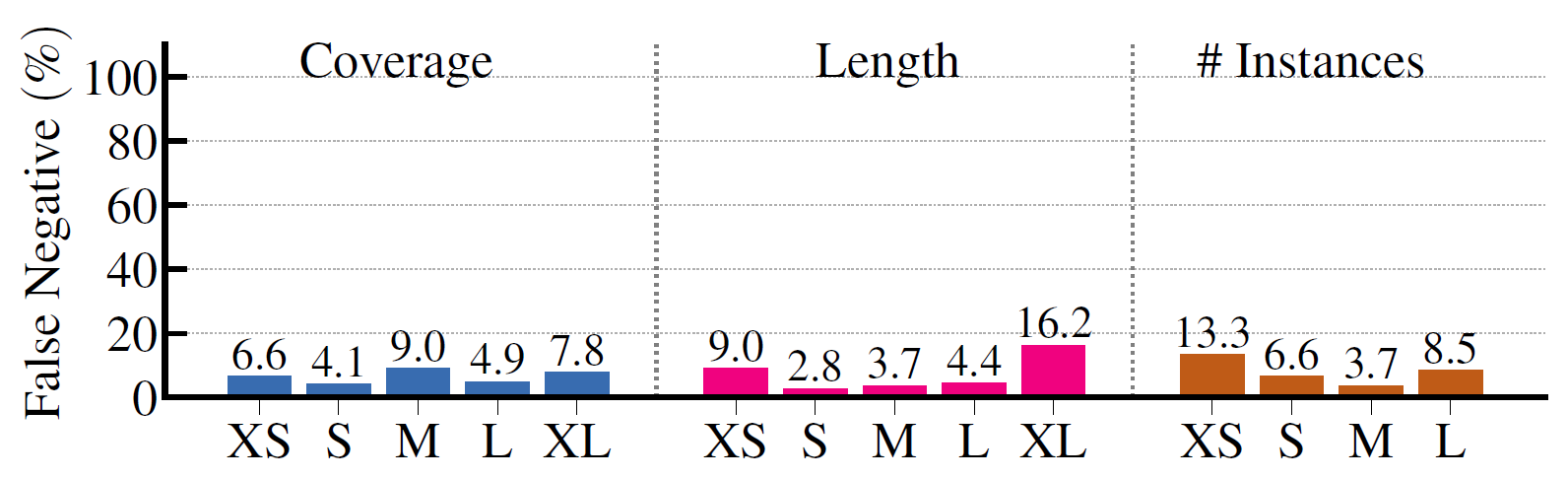}
        \caption{ActionFormer \cite{zhang2022actionformer} (Fully-supervised).}
        \label{fig:ActionFormer_FN}
    \end{subfigure}
    \hfill
    \begin{subfigure}[b]{0.8\linewidth}
        \includegraphics[width=\linewidth]{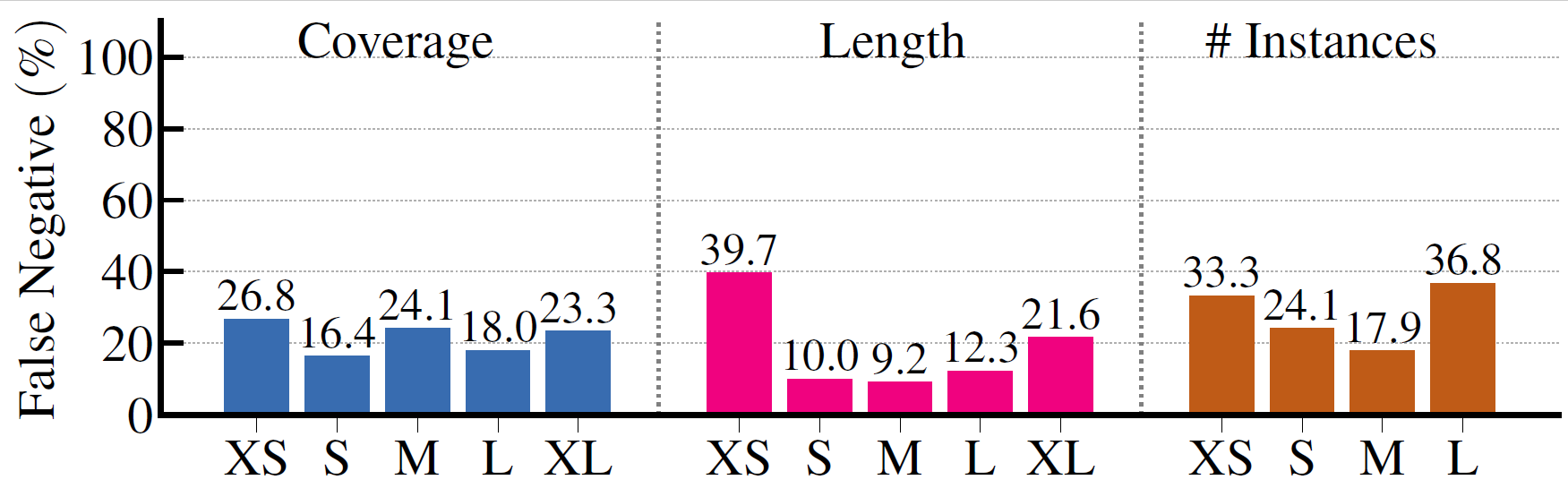}
        \caption{Our POTLoc model (Point-supervised).}
        \label{fig:POTLoc_FN}
    \end{subfigure}
    \hfill
    \begin{subfigure}[b]{0.8\linewidth}
        \includegraphics[width=\linewidth]{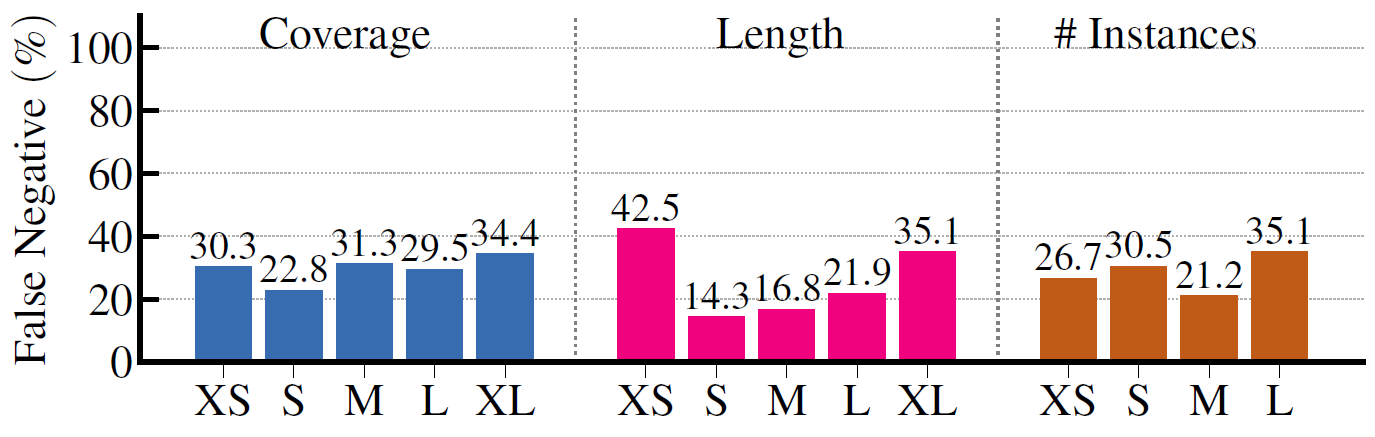}
        \caption{Our base model (Point-supervised).}
        \label{fig:base_FN}
    \end{subfigure}
    \caption{False negative profiling of ActionFormer \cite{zhang2022actionformer} (fully-supervised), POTLoc (point-supervised) and the base model (point-supervised) on THUMOS14 using DETAD \cite{alwassel2018diagnosing}.}
    \label{fig:combined_FN}
\end{figure}

\begin{figure*}[t!]
    \centering
    \begin{subfigure}[b]{\textwidth}
        \centering
        \includegraphics[width=\textwidth]{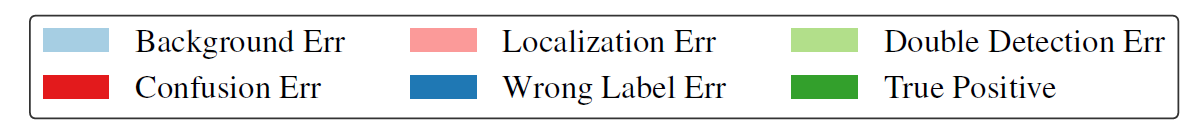}
    \end{subfigure}
    \vspace{1em}
    \begin{subfigure}[b]{0.32\textwidth}
        \centering
        \includegraphics[width=\textwidth]{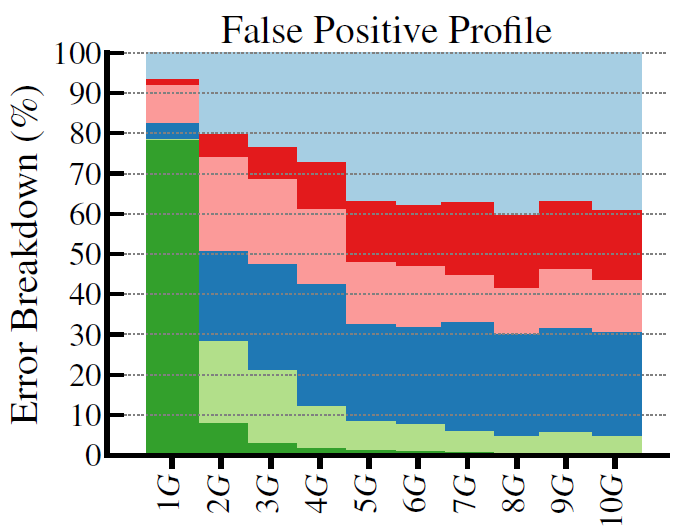}
        \caption{ActionFormer\cite{zhang2022actionformer}(Supervised)}
    \end{subfigure}
    \hfill 
    \begin{subfigure}[b]{0.32\textwidth}
        \centering
        \includegraphics[width=\textwidth]{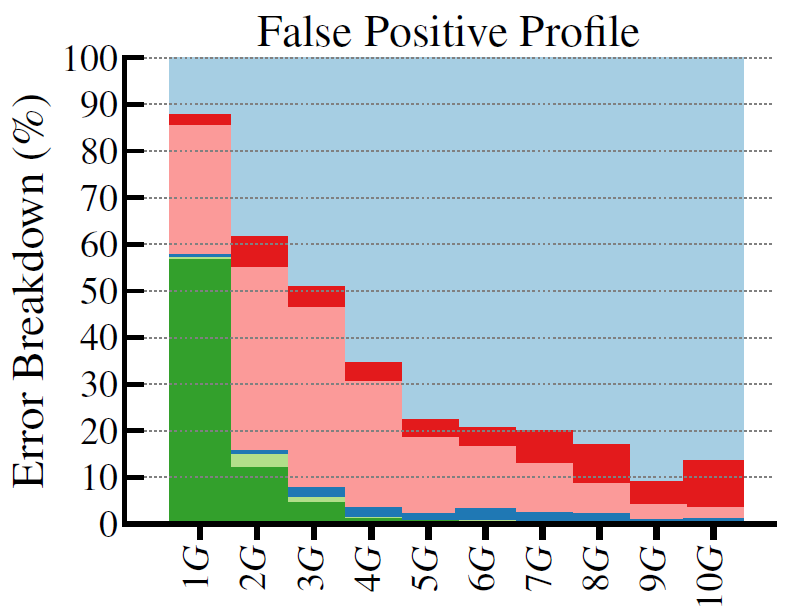}
        \caption{POTLoc (Point-supervised)}
    \end{subfigure}
    \hfill 
    \begin{subfigure}[b]{0.32\textwidth}
        \centering
        \includegraphics[width=\textwidth]{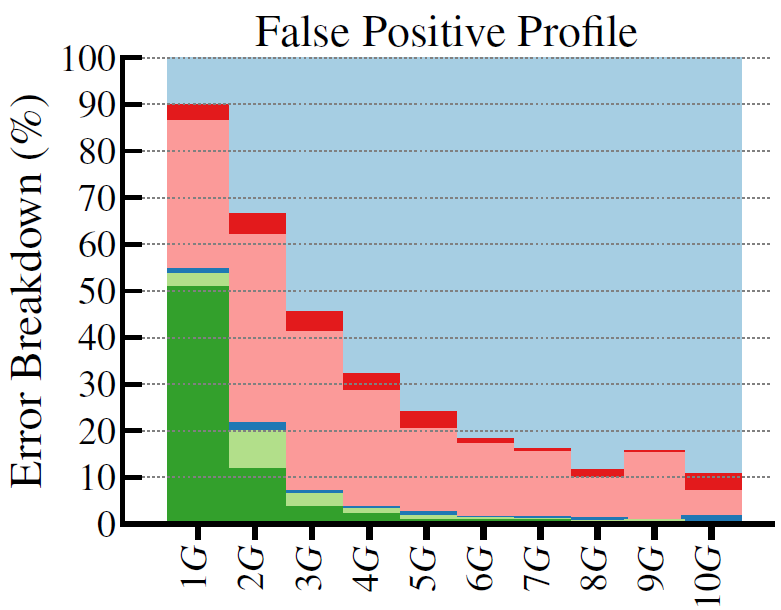}
        \caption{Base model (Point-supervised)}
    \end{subfigure}
    \caption{False positive (FP) profiling of ActionFormer \cite{zhang2022actionformer} (fully-supervised), POTLoc (point-supervised) and base model (point-supervised) on THUMOS14 using DETAD \cite{alwassel2018diagnosing}.}
    \label{fig:FP_profiles}
\end{figure*}

\subsection{Temporal Action Detection Error Analysis}

DETAD \cite{alwassel2018diagnosing} is empolyed for analyzing false negatives (Fig.~\ref{fig:combined_FN}) and false positives (Fig.~\ref{fig:FP_profiles}) of POTLoc in comparison with the base model and a fully-supervised method (ActionFormer\cite{zhang2022actionformer}).

\textbf{False Negative Analysis}. Fig.~\ref{fig:combined_FN} illustrates the false negative (FN) profiling across various coverages, lengths, and number of instances. Part (b) of Fig.~\ref{fig:combined_FN} displays the FN profiling of POTLoc. The figure reveals that higher false negative rates are associated with action instances characterized by: (1) extremely short (Coverage (XS)) or extremely long (Coverage (XL)) durations relative to the video length , (2) actions of very short or very long lengths (Length (XS) or Length (XL)), and (3) videos containing very small ($\#$Instances (XS)) or large number of action instances ($\#$Instances (L)). Furthermore, Fig.~\ref{fig:combined_FN} demonstrates that POTLoc (part b) reduces the false negative (FN) rate compared to the base model (part c) in most cases. FN profiling of ActionFormer\cite{zhang2022actionformer} is provided (part a) which has much lower false negative rate compared with POTLoc because of access to the annotation of action boundaries.

\textbf{False Positive Analysis}. Fig.~\ref{fig:FP_profiles} presents a detailed categorization of false positive errors and summarizes their distribution. In comparison with Actionformer (part a), the majority of false positive errors in POTLoc (part b) stem from background errors. This occurs because POTLoc lacks access to precise action boundaries. Therefore, background snippets close to action boundaries may be erroneously detected as actions, resulting in false positives. Moreover, the false positive profiling of POTLoc (part b) is compared against the base model (part c). POTLoc detects more true positive instances and exhibits fewer localization and confusion errors which confirms the effectiveness of POTLoc compared to the base model.

\subsection{Qualitative Results}

\begin{figure*}[t!]
    \centering
    \begin{subfigure}[b]{\linewidth}
        \centering
        \includegraphics[width=\linewidth]{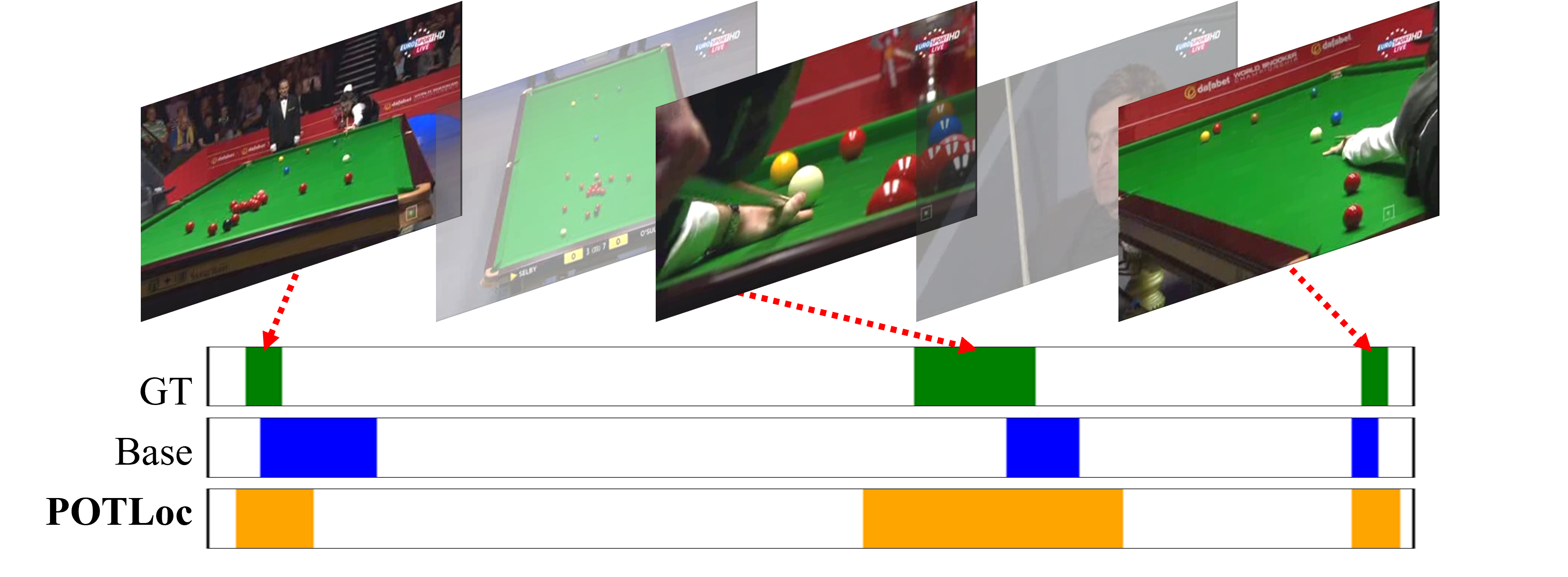}
        \caption{Action ``Billiards".}
    \end{subfigure}
    \hfill 
    \begin{subfigure}[b]{\linewidth}
        \centering
        \includegraphics[width=\linewidth]{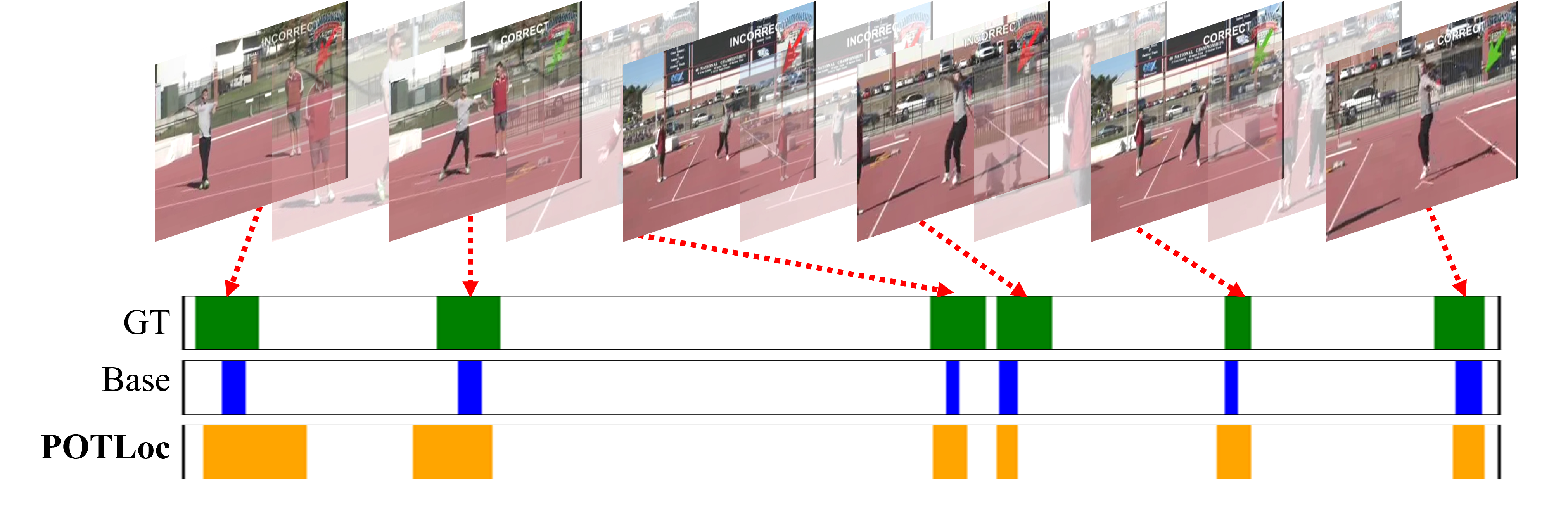}
        \caption{Action ``Javelin Throw".}
    \end{subfigure}
    \hfill 
    \begin{subfigure}[b]{\linewidth}
        \centering
        \includegraphics[width=0.95\linewidth]{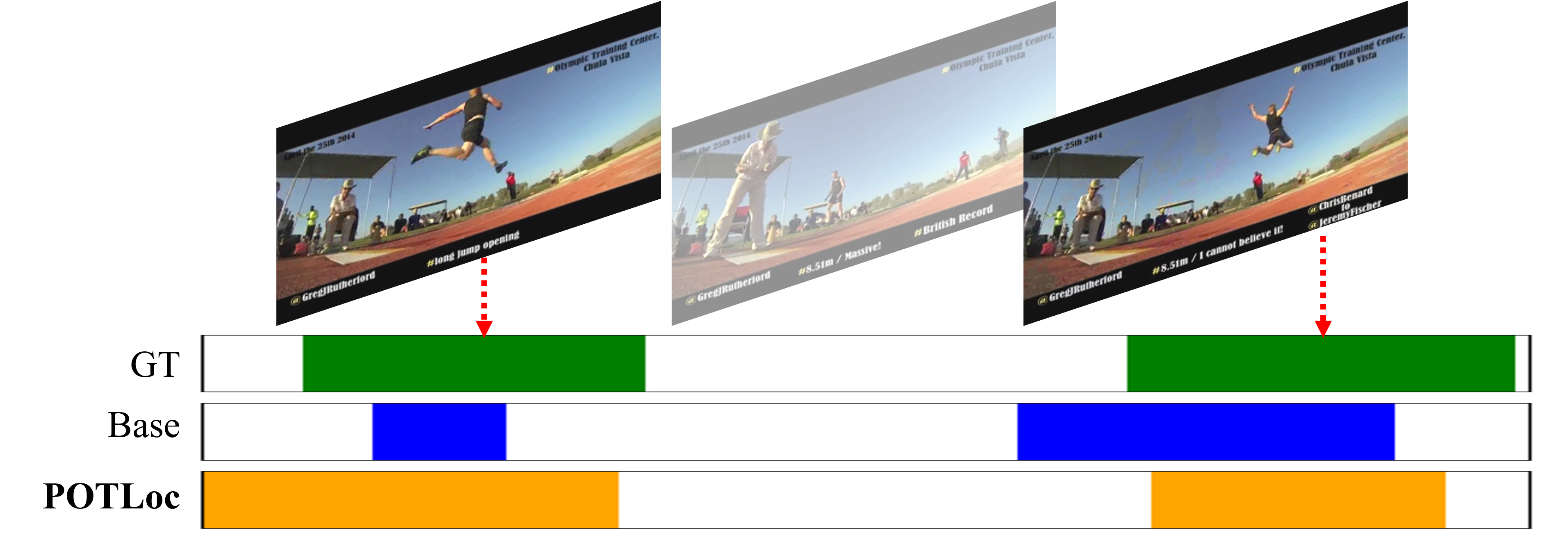}
        \caption{Action ``Long Jump".}
    \end{subfigure}
    \caption{Qualitative results on THUMOS'14. The ground-truth instances are highlighted in green. The detection results are displayed from: (1) the base model supervised with point-level annotations (blue), and (2) our POTLoc framework (orange). Transparent frames represent background frames.}
    \label{fig:vis_results}
\end{figure*}

Fig.~\ref{fig:vis_results} presents the qualitative results of our model in comparison with the base model. POTLoc addresses various types of errors in the base model such as incompleteness and misalignment. In some cases, POTLoc successfully detects complete action proposals, whereas the base model tends to detect fragmented and disconnected segments of action instances. However, as a limitation of POTLoc, in some cases the predicted proposals are over-completed (expanded beyond the action boundaries).



\section{Conclusion}

We have proposed a novel point-supervised framework, POTLoc, that employs a self-training scheme to effectively learn action dynamics. A unique strategy is formulated for pseudo-label generation, which refines action proposals generated from the base model, thus offering supplemental supervisory signals. The effectiveness of the proposed approach for generating and sampling pseudo-labels is confirmed through our experiments. We further elucidated how the transformer and the feature pyramid network utilize the guidance from pseudo-labels to accurately model continuous action structures and handle actions of various durations. POTLoc outperforms the state-of-the-art methods on THUMOS'14 dataset and ActivityNet-v1.2.

\section{Acknowledgment}
This material is based upon work supported by the National Science Foundation under award number 2041307.

 \bibliographystyle{elsarticle-num} 
 \bibliography{cas-refs}





\end{document}